\documentclass[letterpaper, 10 pt, conference]{ieeeconf}

\IEEEoverridecommandlockouts

\usepackage{etextools}
\usepackage{amssymb}
\usepackage{float}
\usepackage{makeidx}
\usepackage{amsmath}
\usepackage{bbm}
\usepackage{epsfig}
\usepackage{epsf}
\usepackage{psfrag}
\usepackage{verbatim}
\usepackage{color}
\usepackage{multirow}
\usepackage{tabularx}
\usepackage{booktabs}
\usepackage[tight,footnotesize]{subfigure}
\usepackage{array}
\usepackage{soul}
\usepackage{footnote}
\usepackage{lipsum}
\usepackage{cite}
\usepackage{dblfloatfix}

\usepackage{color, colortbl}
\usepackage[colorlinks,bookmarksnumbered,citecolor=orange,urlcolor=orange]{hyperref}
\usepackage{graphicx}
\graphicspath{{Figures}}
\DeclareGraphicsExtensions{.pdf,.png}
\usepackage{bigstrut}
\usepackage[english]{babel}

\usepackage{csquotes}

\usepackage[T1]{fontenc}
\usepackage{algorithm, setspace}
\usepackage{algpseudocode}
\usepackage{url}
\usepackage{multirow}
\usepackage{float}
\usepackage{xcolor}
\usepackage{hyperref}
 \hypersetup{
     colorlinks=true,
     linkcolor=orange,
     filecolor=orange,
     citecolor=orange,      
     urlcolor=orange,
     }

\newcommand{\p}[1]{\smallskip \noindent \textbf{{#1}.}}
\newcommand{\eq}[1]{Equation~(\ref{eq:#1})}
\newcommand{\fig}[1]{Figure~\ref{fig:#1}}

\usepackage{balance}

\title{\LARGE
\vspace{-0.6em}
Combining Performance and Passivity in \\ Linear Control of Series Elastic Actuators
\vspace{-0.5em}
}

\author{Shaunak A. Mehta and Dylan P. Losey
\thanks{This work was supported in part by NSF Grant \#2205241. \newline
The authors are members of the Collaborative Robotics Lab (\href{https://collab.me.vt.edu/}{Collab}), Dept. of Mechanical Engineering, Virginia Tech, Blacksburg, VA 24061.
\newline
{Corresponding author's email: \texttt{mehtashaunak@vt.edu}
}
}
\vspace{-0.5em}
}
\begin{document}
\maketitle

%%%%%%%%%%%%%%%%%%%%%%%%%%%%%%%%%%%%%%%%%%%%%%%%%%%%%%%%%%%%%%%%%%%%%%%%%%%%%%%%
\begin{abstract}
When humans physically interact with robots, we need the robots to be both safe and performant. 
Series elastic actuators (SEAs) fundamentally advance safety by introducing compliant actuation. 
On the one hand, adding a spring mitigates the impact of accidental collisions between human and robot; but on the other hand, this spring introduces oscillations and fundamentally decreases the robot's ability to perform precise, accurate motions.
So how should we trade off between physical safety and performance?
In this paper, we enumerate the different linear control and mechanical configurations for series elastic actuators, and explore how each choice affects the rendered compliance, passivity, and tracking performance.
While prior works focus on load side control, we find that actuator side control has significant benefits.
Indeed, simple PD controllers on the actuator side allow for a much wider range of control gains that maintain safety, and combining these with a damper in the elastic transmission yields high performance.
Our simulations and real world experiments suggest that, by designing a system with low physical stiffness and high controller gains, this solution enables accurate performance while also ensuring user safety during collisions.
\end{abstract}

% \smallskip

\vspace{-0.25em}
%%%%%%%%%%%%%%%%%%%%%%%%%%%%%%%%%%%%%%%%%%%%%%%%%%%%%%%%%%%%%%%%%%%%%%%%%%%%%%%%
\section{Introduction}

Robots are becoming increasingly common in the daily lives of humans, moving from controlled and isolated settings in factory floors to collaborative environments. 
When these robots work alongside human users, physical interaction between humans and robots is inevitable. 
This introduces a question of safety for the humans working alongside these robots. 
For example, if a robot is performing a high-speed task and the human intervenes --- intentionally or unintentionally --- the impact with the human may cause serious injuries \cite{haddadin2016physical}. 
Hence, we need to ensure safety of the human and performance of the robot.

Series Elastic Actuators (SEAs) provide a mechanical solution for safety during human-robot interactions.
SEAs are formed by introducing a compliant element between the actuator and the output of the system \cite{pratt1995series} (see Figure \ref{fig:front}). 
When a robotic system with such compliant elements collides with a user, the compliant element softens the impact.
That is, instead of being impacted by the rigid body of the robot, the users feel the compliance of the spring in the system. 
As opposed to safe control approaches that operate at a fixed frequency and may fail to act promptly given an impact, SEAs advance the fundamental safety of the system in a way that control-based approaches cannot achieve. 
However, the addition of a compliant element introduces oscillations in high-inertia robots making control of the system more complex \cite{hurst2004series}. 
Recent works have explored different solutions to ensure performance after the introduction of this compliant element. 
Some approaches focus on the mechanical aspect of the system, exploring solutions involving additional damping \cite{shardyko2020development} and non-linear stiffness \cite{wolf2015variable}. 
Other approaches focus on control mechanisms for the series elastic actuators \cite{sharkawy2022human}.

Across these prior works, however, we consistently find the trade-off between increasing compliance and performance. 
Intuitively --- the softer the spring, the safer the system during impact. 
But softer springs also inherently limit the range of control gains we can choose \cite{kenanoglu2023fundamental}, and add oscillations that prevent accurate trajectory tracking \cite{hurst2004series}.
We want to answer the question: how can we ensure both safety and accuracy of the system when interacting with humans in the real world? 
While the mechanical and controls solutions have their drawbacks individually, our insight is that:
\begin{center}
\textit{SEAs can ensure safety and performance of the system by combining physical damping with actuator-side control.}
\end{center}
In this paper we systematically evaluate mechanical and linear control solutions for SEAs. 
We explore each approach along three axis: 1) the rendered compliance of the system, 2) the stability of the system when coupled with a human, and 3) the performance of the system in terms of position tracking. 
Building on the findings from this theoretical analysis and leveraging our insight, we propose a solution that combines mechanical damping and actuator-side control for series elastic actuators. 
This design and control combination guarantees compliance and coupled stability during human-robot interaction, while also allowing for high gains and accurate tracking performance.

Overall, we make the following contributions:\\
\p{Passivity Analysis of Design and Control Strategies} 
We perform passivity analysis to get the bounds on physical and controller parameters under which the different design and linear control architectures of SEAs maintain coupled stability. 
We then consider the systems that are passive for a wide range of control parameters and analyze the behavior of each system along the axes of compliance and performance.

\p{Combining Mechanical and Control Approaches}
Building on our analysis of the different design and control strategies, we propose a system that combines damped SEA design with actuator-side control. 
We then derive the conditions for passivity of this system and analyze its performance and safety. 
This system allows for a more accurate control over the load position while also ensuring safety for a wide range of impact frequencies. 

\p{Simulated and Real-World Experiments}
We compare the performance of our proposed approach to different controllers in simulated as well as real world experiments. 
The simulated tests show the importance of physical damping for the performance of the system, while the real world experiments highlight its accuracy in position tracking and low perceived stiffness when interacting with humans.

% \vspace{-0.35em}
\section{Related Work} \label{sec:related}

\noindent \textbf{Compliant Actuation.} 
Recent works have focused on making robotic systems safe during interaction with humans \cite{lasota2017survey,de2008atlas,li2024safe}. 
Some approaches here include safety filters \cite{sharkawy2022human}, motion planning \cite{hu2022sharp} and human-behavior prediction \cite{liu2024intention}. 
Compliant actuation has also been explored to make the robotic systems \textit{physically safe by design} \cite{van2009compliant}. 
Traditionally, SEAs introduce a compliant element having a fixed stiffness between the actuator and the output of the system \cite{pratt1995series}. 
During collisions, these compliant elements can soften the impact, enabling safe interactions with humans. 
However, due to the introduction of this compliant element, the performance of the system may be compromised \cite{hurst2004series}. 
Some works have explored variable stiffness actuators (VSAs) that change the stiffness of the compliant elements by introducing an additional physical system, generally in the form of motors \cite{wolf2015variable}. 
While these systems can change the mechanical stiffness of the system (i.e., modulate the physical spring stiffness), this change in stiffness requires control decisions and cannot be achieved instantaneously in response to collisions.
 
\p{Stable Control of SEAs}
Given these challenges for mechanical solutions, we next turn to the different controllers used for compliant systems. 
Control architectures specific to compliant mechanisms can tackle the trade-off between safety and performance \cite{abu2020variable}.
Cascaded controllers have been used for force control of compliant actuators by breaking the controller down into position and velocity components \cite{tagliamonte2014rendering,vallery2007passive}. 
Similarly, velocity sourced impedance controllers (VSIC) have been explored to directly control the output velocity \cite{mengilli2021passivity,tosun2020necessary}.
Unfortunately, these approaches introduce fundamental limitations on the physical and control parameters necessary to maintain stability during human-robot interaction.\cite{kenanoglu2023fundamental,tagliamonte2014passivity}. 
These constraints may limit the system accuracy or compromise the safety of the system during interactions with humans. 
Some works have tried incorporating physical damping to relax the controller constraints \cite{calanca2017impedance,ghidini2020robust,shardyko2020development,kenanoglu2024passive,monteleone2022damping}. 
Our work is most similar to \cite{shardyko2021series}, which adds active virtual damping into the linear control of the system, and \cite{kenanoglu2025effect}, which derives relaxed stability bounds for VSIC in the presence of physical damping. 
These approaches minimize some of the constraints on the physical and controller parameters of the system.
However, since they rely on the load-side control, their controller gains are still bounded by the physical stiffness of the compliant element.
Contrary to these approaches, we instead propose an alternative perspective using actuator-side control, and demonstrate that --- with the addition of damping into the compliant system --- we can outperform load-side control in terms of both safety and performance.
\section{Problem Statement} \label{sec:problem}

We consider scenarios with a 1-degree of freedom (DoF) SEA and linear control. 
Our goal is to design a hybrid system with mechanical and control components that is precise and stable when coupled to human user, ensuring safety and accuracy. 
In this section we will discuss the preliminary details regarding the design of the series elastic actuators and define what it means for a SEA to be passive. 

\begin{figure}[t]
	\begin{center}
		\includegraphics[width=1.0\columnwidth]{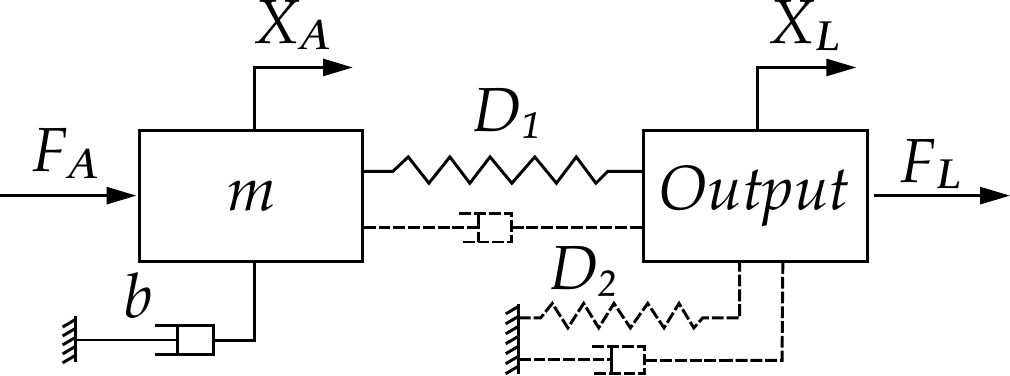}
		\vspace{-0.5em}
		\caption{Schematic of a 1-DoF series elastic actuator (SEA). The actuator with mass $m$ and internal damping $b$ is connected to the output via a compliant element $D_1$. 
        In general, the transmission could include either a pure spring or a spring and damper. By default, the load side is not connected to the ground. But we explore generalizations with a compliant element $D_2$ that again could consist of a spring and/or damper. $F_A$ and $F_L$ are the forces acting on the motor and the output, and $X_A$ and $X_L$ denote the position of the actuator and the output. We want to design a system that can be enable accurate tracking of the load position $X_L$, while also rendering a low stiffness at the output during interactions with humans.}
		\label{fig:front}
	\end{center}
 	\vspace{-2.0em}
\end{figure}

\p{Series Elastic Actuators}
A general design for 1-DoF damped series elastic actuator, consistent with prior works, is shown in Figure \ref{fig:front} \cite{monteleone2022damping,kenanoglu2024passive}. 
In this figure the actuator with mass $m$ and damping $b$ is connected to a load $L$ via a compliant element $K$. 
More generally, the load may also be connected to the ground via a second compliant element $D_2$. 
The actuator and load positions are given by $X_A$ and $X_L$, respectively. This design of series elastic actuators is acted upon by two forces, namely, the force applied by the actuator to move the motor ($F_A$) and the force applied by the human on the load ($F_L$). 
The force $F_L$ is considered to be positive when the spring is in tension. The equations of motion for this system are:
\begin{equation} \label{eq:p1}
    f_A + f_L = m\ddot{x}_A(t) + b\ddot{x}_A(t)
\end{equation}
\begin{equation}\label{eq:p2}
    f_L = D_1(t)(x_L(t) - x_A(t)) + D_2(t)x_L(t)
\end{equation}
Let $\theta_A$ and $\theta_L$ be the desired positions of the actuator and the load, and let $C_A$ and $C_L$ be the controllers on the actuator and load side that move the system towards the desired positions. 
For example, $C_A$ could be a proportional controller that regulates the forces of the actuator based on the displacement between $\theta_A$ and $x_A$. 
The combined force output of the actuator, based on these controllers, is:
\begin{equation}\label{eq:p3}
    f_A = C_A(t)(\theta_A - x_A(t)) + C_L(t)(\theta_L - x_L(t))
\end{equation}
Since the user interacts with the system on the load side, the stiffness perceived by the user ($k_{obs}$) can be found by analyzing the force and displacement of the load during an interaction.
Specifically, the rendered stiffness follows $F_L = k_{obs}X_L$. 
Combining Equations (\ref{eq:p1})--(\ref{eq:p3}), we can write the perceived stiffness of the system in Laplace domain as:
\begin{equation} \label{eq:p4}
    \dfrac{F_L}{X_L} = \dfrac{(D_1(s) + D_2(s))(\alpha+C_A(s)) + D_1(s)C_L(s)}{\alpha + C_A(s) + D_1(s)}
\end{equation}
where $\alpha = ms^2 + bs$. 
We want to design a system that, on impact with an external force, remains compliant. 
In other words, we want this perceived stiffness ($F_L/X_L$) to be low, so that human experiences a softened impact.

\p{Passivity} 
A system is considered to be passive if it does not generate energy. 
That is, the energy output by the system is less than or equal to the energy that is input to drive the system \cite{colgate1988robust}. 
Studying the passivity of a system can provide us with the range of allowable control and physical parameters that ensure the system will remain stable for all operating conditions when coupled to any other passive system. 
Thus, passivity provides us with a sufficient condition for safe interaction when we try to balance the trade-off between safety and performance. 
For our case of SEAs, the interaction passivity of the system can determined by analyzing the impedance transfer function \cite{losey2017effects,tosun2020necessary}:
\begin{equation} \label{eq:p5}
    Z_L(s) = \dfrac{1}{s} \cdot \dfrac{F_L}{X_L}
\end{equation}
Following \cite{colgate1988robust}, a system with a transfer function $Z_L(s)$ is considered to be passive if $Z_L(s)$ is positive real. That is, if the following three conditions are satisfied: 
\begin{itemize}
    \item the poles of $Z_L(s)$ on the imaginary axis are simple.
    \item $Z_L(s)$ is stable.
    \item $Re(Z_L(s)) \geq 0 \ \forall \ \omega \in \mathbb{R}$ where $j\omega$ is not a pole.
\end{itemize}
The first two conditions are always satisfied since the only pole of the system on the imaginary axis is at the origin and the coefficients of the denominator are always positive \cite{ogata2010modern}. 
Thus, for ensuring the passivity of the system, we require that the third condition is satisfied. 
The real part of $Z_L(s)$ can be given as:
\begin{equation} \label{eq:p6}
\begin{split}
    Re(Z_L(j\omega)) &= Re\bigg(\dfrac{Num(j\omega)Den(-j\omega)}{Den(j\omega)Den(-j\omega)}\bigg)
\end{split}
\end{equation}
where $Num(\cdot)$ and $Den(\cdot)$ represent the numerator and denominator of $Z_L(\cdot)$. 
Since the denominator of Equation \ref{eq:p6} is always non negative, we get the following condition for SEA passivity:
\begin{equation}\label{eq:p7}
Re(Num(j\omega)Den(-j\omega)) \geq 0
\end{equation}
We will leverage passivity analysis to examine the performance and safety of a system with a given controller and mechanical design. 
For example, how much can we crank up the controller gains or decrease the physical stiffness of the system while ensuring that the system remains passive (i.e., coupled stable)? 
In the next section we will explore how different mechanical and control choices affect the passivity and perceived stiffness of the system.

\section{Design and Control Analysis of SEAs } \label{sec:theory}
In the previous section we provided an overview of a general design for series elastic actuators and outlined its conditions for passivity.
In this section, we now evaluate how different mechanical and controller designs can affect passivity, rendered compliance, and performance.
Specifically, in Section~\ref{ss:t1} we explore the effect that physical changes --- i.e., additional springs and dampers --- have on the system, and in Section~\ref{ss:t2} we study how different linear control strategies change the performance of the system. 
Note that we only consider the linear controllers shown in Section \ref{sec:problem}, where $C_A$ and $C_L$ can be arbitrary transfer functions. Finally, in Section~\ref{ss:t3}, we combine the solutions from design and control perspectives and propose a system that is passive, compliant, and high-performance.

\subsection{Mechanical Approaches} \label{ss:t1}

Here we evaluate how different configurations of springs and dampers affect series elastic actuators. 
Consistent with prior works, by default we use \textit{load-side proportional control}. 
That is, we set $C_A(s) = 0$ and $C_L(s) = k_d$, where $k_d$ is the virtual stiffness of the controller. 
Substituting these controller parameters in Equation \ref{eq:p4}, the stiffness transfer function simplifies to:
\begin{equation}\label{eq:m1}
    \dfrac{F_L}{X_L} = \dfrac{(D_1(s) + D_2(s))\alpha + D_1(s)k_d)}{\alpha + D_1(s)}
\end{equation}
For each of the spring and damper configurations we consider in this section, we define the compliant element of the SEA transmission, $D_1(s)$, and any connection between the load and the ground, $D_2(s)$. 
We then analyze the passivity of the resulting system and report the stiffness as a function of frequency. This analysis is summarized in Table \ref{tab:mech}.

\begin{table*}[]
\centering
\caption{Passivity and compliance analysis of different mechanical configurations of SEAs with load side proportional control} \label{tab:mech}
\scriptsize
\vspace{-1.0em}
\begin{tabular}{@{}c@{\hspace{0.5em}}c@{\hspace{1pt}}c@{}c@{\hspace{1pt}}c@{\hspace{5pt}}c@{}}
\toprule
Method & Parameters & Transfer Function & Passivity Condition & Frequency Response \\ \midrule
   \begin{tabular}{c@{\hspace{2pt}}c} \rotatebox[origin=c]{90}{Pure} & \rotatebox[origin=c]{90}{Spring} \end{tabular}    & 
   % \begin{tabular}{c}$D_1(s) = k_1, D_2(s) = 0$ \\ $C_A(s) = 0$, $C_L(s) = k_d$\end{tabular}    
   \begin{tabular}{c}$D_1(s) = k_1$ \\ $D_2(s) = 0$\end{tabular}    
   &     $\dfrac{k_1(\alpha + k_d)}{\alpha + k_1}$               &      $ bk_1(k_1-k_d)\omega^2 \geq 0$              &     \raisebox{-0.4\height}{\includegraphics[width=0.4\columnwidth]{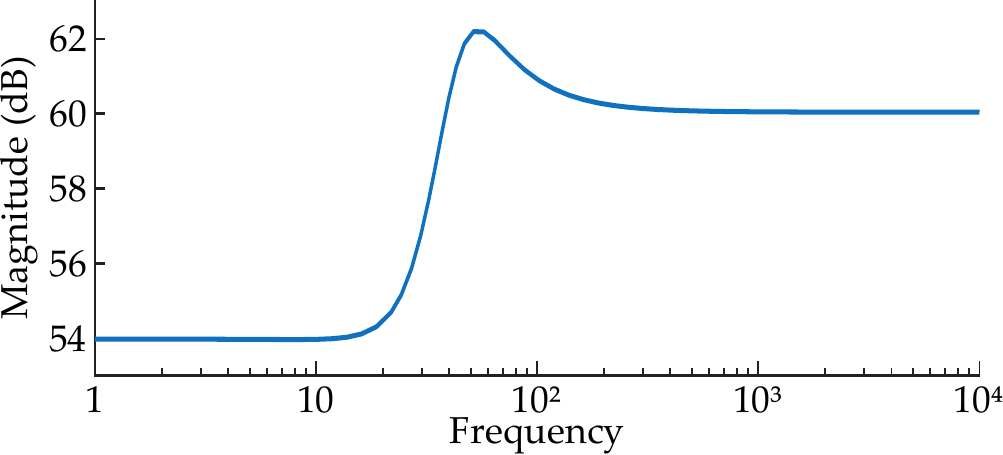}}      \\
   \midrule
   
   \begin{tabular}{c@{\hspace{2pt}}c} \rotatebox[origin=c]{90}{Disjointed} & \rotatebox[origin=c]{90}{Spring-Spring} \end{tabular}    & 
   % \begin{tabular}{c}$D_1(s) = k_1$, $D_2(s) = k_2$ \\ $C_A(s) = 0$, $C_L(s) = k_d$\end{tabular}     
   \begin{tabular}{c}$D_1(s) = k_1$ \\ $D_2(s) = k_2$ \end{tabular}              
   &     $\dfrac{(k_1+k_2)\alpha + k_1k_d}{\alpha + k_1}$               &      $ bk_1(k_1+k_2-k_d)\omega^2 \geq 0$              &     \raisebox{-0.4\height}{\includegraphics[width=0.4\columnwidth]{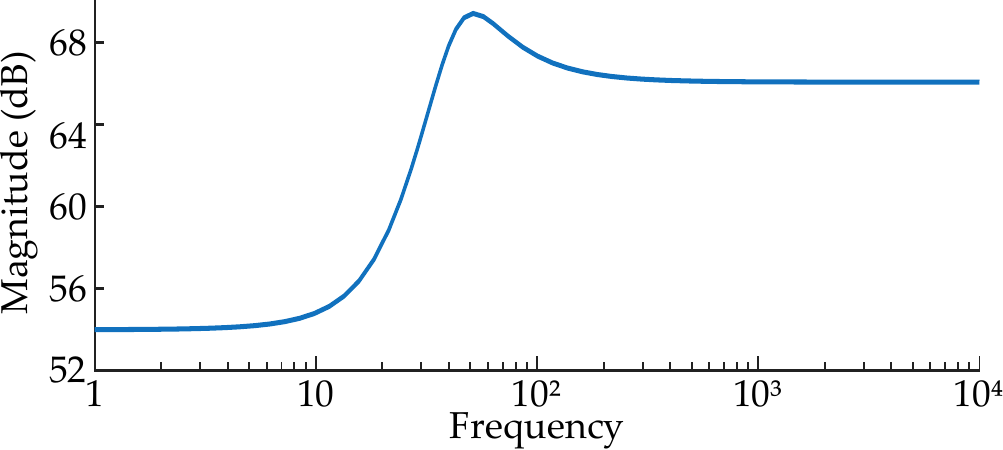}}      \\
   \midrule
   
   \begin{tabular}{c@{\hspace{2pt}}c} \rotatebox[origin=c]{90}{Parallel} & \rotatebox[origin=c]{90}{Spring-Damper} \end{tabular}    & 
   % \begin{tabular}{c}$D_1(s) = k_1 + b_1s$, $D_2(s) = 0$ \\ $C_A(s) = 0$, $C_L(s) = k_d$\end{tabular}              
   \begin{tabular}{c}$D_1(s) = k_1 + b_1s$\\ $D_2(s) = 0$\end{tabular}              
   &     $\dfrac{(k_1+b_1s)(\alpha + k_d)}{\alpha + k_1 + b_1s}$               &      \begin{tabular}{c@{}c}{\begin{tabular}{c}$(b^2 + b\cdot b_1 - m\cdot k_d)b_1\omega^4$ \\ $+ (k_1 - k_d)bk_1\omega^2$\end{tabular}} & $\geq 0$\end{tabular}             &     \raisebox{-0.4\height}{\includegraphics[width=0.4\columnwidth]{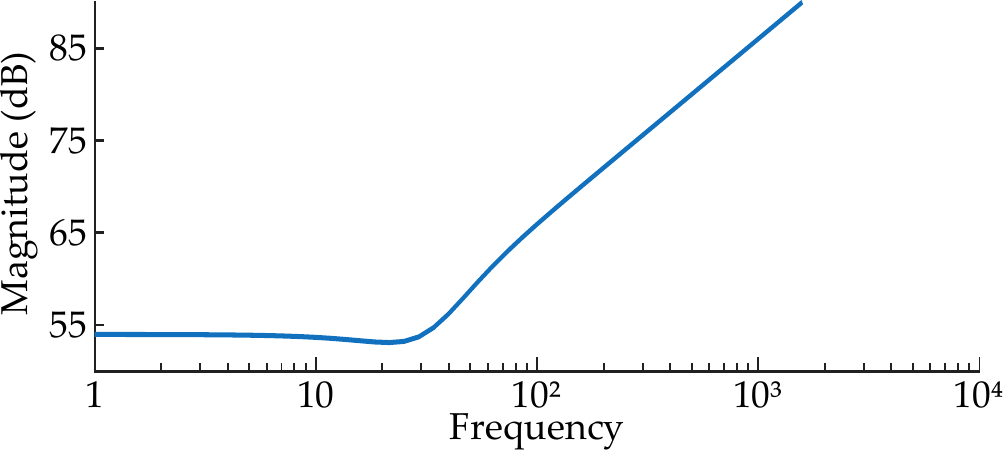}} \\
   \midrule
   
   \begin{tabular}{c@{\hspace{2pt}}c} \rotatebox[origin=c]{90}{Disjointed} & \rotatebox[origin=c]{90}{Spring-Damper} \end{tabular}    & 
   % \begin{tabular}{c}$D_1(s) = k_1$, $D_2(s) = b_2s$ \\ $C_A(s) = 0$, $C_L(s) = k_d$\end{tabular}   
   \begin{tabular}{c}$D_1(s) = k_1$\\ $D_2(s) = b_2s$ \end{tabular} 
   &     $\dfrac{(k_1+b_2s)\alpha + k_1k_d}{\alpha + k_1}$               &      \begin{tabular}{c@{}c}{\begin{tabular}{c}$ mb_2\omega^6 + b_2(b^2 - mk_1)\omega^4$ \\ $+ bk_1(k_1 - k_d)\omega^2$\end{tabular}} & $\geq 0$\end{tabular}             &     \raisebox{-0.4\height}{\includegraphics[width=0.4\columnwidth]{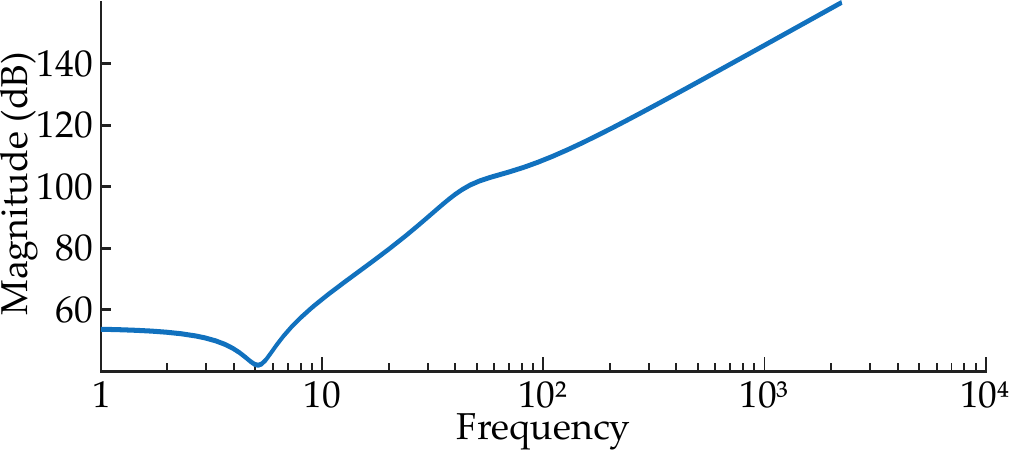}} \\
       \bottomrule
\end{tabular}
\vspace{-1.em}
\end{table*}

\subsubsection{Pure Spring} 
We first consider a system where the load is connected to the actuator via a pure spring $D_1 = k_1$ and there is no compliant element between the load and the ground $D_2 = 0$. 
Substituting these values into \eq{m1}, we obtain the stiffness transfer function and passivity condition shown in Row 1 of Table \ref{tab:mech}.
This suggests that the system will be always be passive if the controller stiffness is smaller than or equal to the physical stiffness of the spring, i.e., $k_d \leq k_1$. 
The frequency response for this controller demonstrates that --- as the frequency of control increases --- the perceived stiffness of the system goes to the mechanical spring stiffness $k_1$. 
By designing a system with a low stiffness $k_1$ we can ensure higher compliance. 
However this also limits the accuracy of the system due to the constraints on the controller stiffness, thus making its use in robotic applications challenging.

\subsubsection{Disjointed Spring-Spring}
We next consider a system with two springs, one in the transmission ($D_1(s) = k_1$), and the other connecting the load to the ground ($D_2(s) = k_2$). 
Row 2 in Table \ref{tab:mech} provides the stiffness transfer function and passivity conditions for this case.
The behavior of this system with an external spring connected to the ground, where the condition for passivity can be given as $k_d \leq k_1 + k_2$, is similar to that of the single pure spring connecting the load to the actuator.
This again makes the use of this system difficult in human-robot interaction scenarios where we want to increase gains for high performance.

\subsubsection{Parallel Spring-Damper}
Until now we have considered cases with pure springs, inducing no additional damping in the system. 
However, in the real world, we do not have access to pure springs due to friction in the transmission and the internal damping of the system. 
In this case, we model the system with spring constant $k_1$ and damping $b_1$ ($D_1(s) = b_1s + k_1, D_2(s) = 0$). Substituting this value of $D_1(s)$ in \eq{m1}, we get the stiffness transfer function for a parallel spring-damper system (Row 3 of Table \ref{tab:mech}). 
Leveraging equation \eq{p7}, the passivity conditions suggests that this system will always be passive when $k_d \leq k_1$ and $k_d \leq (b^2 + b\cdot b_1)/m$. 
These conditions, depending on the mechanical stiffness as well as the damping of the system, require that we either have high damping or low controller stiffness.
Looking back at the frequency response of the system in Table \ref{tab:mech}, we observe that the stiffness of the system increases with an increase in the frequency due to the additional damping. 
However, this increase in perceived stiffness may still allow for safe interactions based on the choice of the system parameters ($m, b, k_1, b_1$), thus trading-off between the safety and performance of the system.

\subsubsection{Disjointed Spring-Damper}
Finally, we consider a case of a spring with stiffness $D_1(s) = k_1$ connecting the actuator to the load and a damper with damping constant $D_2(s) = b_2(s)$ attached between the load and the ground. 
The stiffness transfer function for this system and the corresponding passivity conditions are provided in Row 4 of Table \ref{tab:mech}.
Since $m, b, k_d, k_1$ and $b_2$ are positive constants, for this system to always be passive, the following constraints should be satisfied: $k_d \leq k_1$ and $k_1 \leq b^2/m$. 
Since the stiffness of the spring $k_1$ depends on the damping of the system, the system requires the use of a spring with a low spring constant. 
This makes the choice of the control and physical parameters more restrictive as compared to the case of a pure spring --- hence, this configuration is strictly worse than the default structure.

Overall, our analysis for mechanical configurations suggests that --- for SEA systems with load-side proportional control --- the most feasible mechanical design is a pure spring connecting the load to the actuator. 
This design offers compliance for safe interactions and is passive for all control gains below the spring's physical stiffness.
But this restriction on the control gains still fundamentally limits performance \cite{losey2017effects}, and thus we turn to control approaches to try and enhance the system.

\subsection{Control Approaches} \label{ss:t2}
In the previous section we evaluated the effect that different configurations of springs and dampers had on the behavior of a compliant system. 
In this section, we now look at this problem from a controls perspective and explore how different controller configurations affect the tradeoff between performance and safety. 
For the analysis in this section we assume the standard SEA setup with a pure spring ($D_1(s) = k, D_2(s) = 0$). Substituting these values in \eq{p4}, we get the stiffness transfer function:
\begin{equation} \label{eq:m10}
    \dfrac{F_L}{X_L} = \dfrac{k_1(ms^2 + bs + C_A(s) + C_L(s))}{ms^2 + bs + k_1 + C_A(s)}
\end{equation}
We follow a linear control architecture and employ different load-side and actuator-side control structures to evaluate the performance of the system. 
For each controller selection, we provide the stiffness transfer function, perform passivity analysis, and discuss the impact safety of the system; the outcomes are summarized in Table \ref{tab:cont}.

\begin{table*}[htp!]
% \vspace{-em}
\centering
\caption{Passivity and compliance analysis of different controllers for SEAs with a pure spring as the compliant element} \label{tab:cont}
\scriptsize
\vspace{-1.0em}
\begin{tabular}{@{}c@{\hspace{0.5em}}c@{\hspace{1pt}}c@{}c@{\hspace{1pt}}c@{\hspace{5pt}}c@{}}
\toprule
 & Method & Parameters & Transfer Function & Passivity Condition & Frequency Response \\ \midrule
   \multirow{3}{*}[-2.5em]{\rotatebox{90}{Load-Side Control}} &
   \begin{tabular}{c} \rotatebox[origin=c]{90}{Proportional}\end{tabular}    & 
   % \begin{tabular}{c}$K(s) = k_1$, $D(s) = 0$ \\ $C_A(s) = 0$, $C_L(s) = k_d$\end{tabular}
   \begin{tabular}{c} $C_A(s) = 0$ \\ $C_L(s) = k_d$\end{tabular}   &     $\dfrac{k_1(\alpha + k_d)}{\alpha + k_1}$               &      $ bk_1(k_1-k_d\omega^2) \geq 0$              &     \raisebox{-0.4\height}{\includegraphics[width=0.4\columnwidth]{figures/bode_m_s_n.pdf}}      \\
   \cmidrule{2-6}

   &
   \begin{tabular}{c@{\hspace{2pt}}c} \rotatebox[origin=c]{90}{Proportional} & \rotatebox[origin=c]{90}{Derivative} \end{tabular}    & 
   % \begin{tabular}{c}$K(s) = k_1$, $D(s) = k_2$ \\ $C_A(s) = 0$, $C_L(s) = k_d$\end{tabular}   
   \begin{tabular}{c} $C_A(s) = 0$ \\ $C_L(s) = k_d + b_ds$\end{tabular}   
   &     $\dfrac{k_1(\alpha + b_ds + k_d)}{\alpha + k_1}$               &     
   \begin{tabular}{c@{}c}{\begin{tabular}{c}$k_1(bk_1 + b_dk_1 -bk_d)\omega^2$ \\ $- b_dk_1m\omega^4$\end{tabular}} & $\geq 0$\end{tabular}      &     \raisebox{-0.4\height}{\includegraphics[width=0.4\columnwidth]{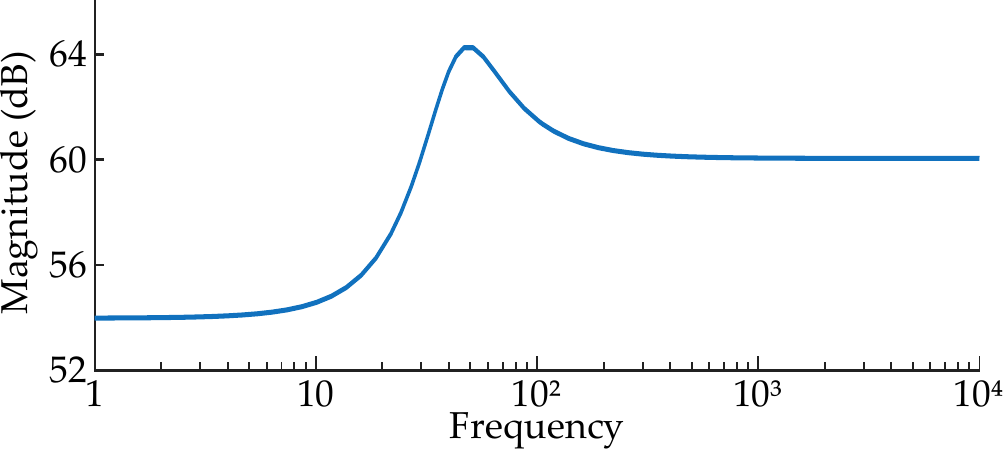}}      \\
   \cmidrule{2-6}

    &
   \begin{tabular}{c@{\hspace{2pt}}c} \rotatebox[origin=c]{90}{Proportional} & \rotatebox[origin=c]{90}{Integral} \end{tabular}    & 
   % \begin{tabular}{c}$K(s) = k_1 + b_1s$, $D(s) = 0$ \\ $C_A(s) = 0$, $C_L(s) = k_d$\end{tabular} 
   \begin{tabular}{c} $C_A(s) = 0$ \\ $C_L(s) = k_d + i_d/s$\end{tabular} 
   &     $\dfrac{k_1(\alpha + k_d + i_d/s)}{\alpha + k_1}$               &      \begin{tabular}{c@{}c}{\begin{tabular}{c}$k_1(bk_1 - bk_d + i_dm)\omega^2$ \\ $-i_dk_1^2$\end{tabular}} & $\geq 0$\end{tabular}             &     \raisebox{-0.4\height}{\includegraphics[width=0.4\columnwidth]{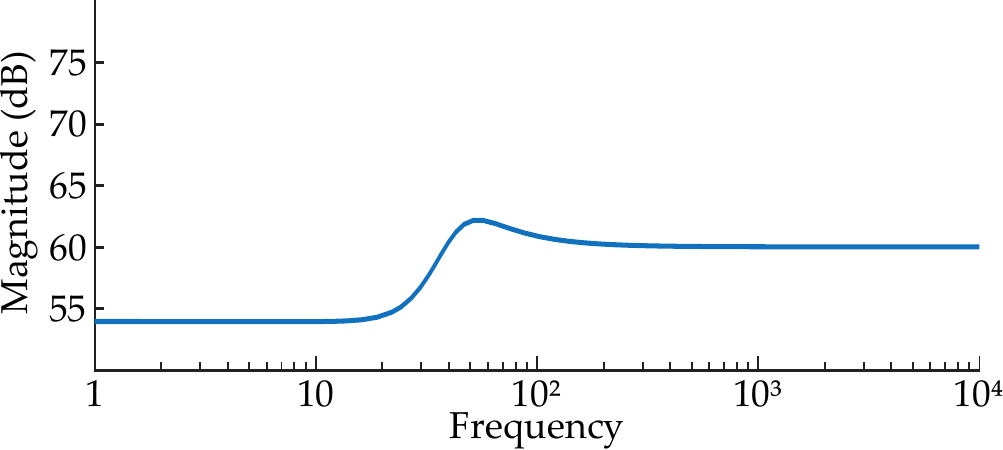}} \\
   \midrule

   \multirow{3}{*}[-2em]{\rotatebox{90}{Actuator-Side Control}}
   &
   \begin{tabular}{c} \rotatebox[origin=c]{90}{Proportional} \end{tabular}    & 
   % \begin{tabular}{c}$K(s) = k_1$, $D(s) = b_2s$ \\ $C_A(s) = 0$, $C_L(s) = k_d$\end{tabular}    
   \begin{tabular}{c} $C_A(s) = k_d$ \\ $C_L(s) = 0$\end{tabular}    
   &     $\dfrac{k_1(\alpha + k_d)}{\alpha + k_1 + k_d}$               &      \begin{tabular}{c@{}c}{\begin{tabular}{c}$bk_1^2\omega^2$\end{tabular}} & $\geq 0$\end{tabular}             &     \raisebox{-0.4\height}{\includegraphics[width=0.4\columnwidth]{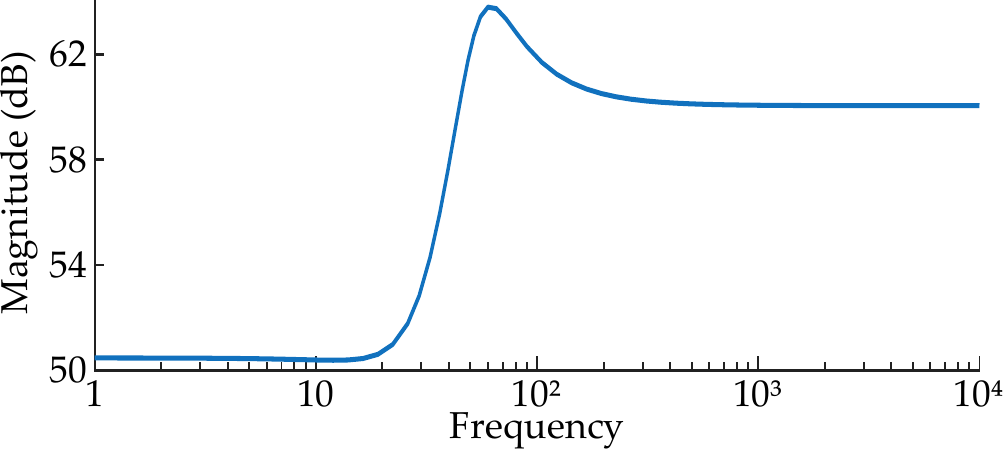}} \\
   \cmidrule{2-6}

   &
   \begin{tabular}{c@{\hspace{2pt}}c} \rotatebox[origin=c]{90}{Proportional} & \rotatebox[origin=c]{90}{Derivative} \end{tabular}    & 
   % \begin{tabular}{c}$K(s) = k_1 + b_1s$, $D(s) = 0$ \\ $C_A(s) = 0$, $C_L(s) = k_d$\end{tabular} 
   \begin{tabular}{c} $C_A(s) = k_d + b_ds$ \\ $C_L(s) = 0$\end{tabular} 
   &     $\dfrac{k_1(\alpha + b_ds + k_d)}{\alpha + b_ds + k_1 + k_d}$               &      \begin{tabular}{c@{}c}{\begin{tabular}{c}$k_1^2(b + b_d)\omega^2$\end{tabular}} & $\geq 0$\end{tabular}             &     \raisebox{-0.4\height}{\includegraphics[width=0.4\columnwidth]{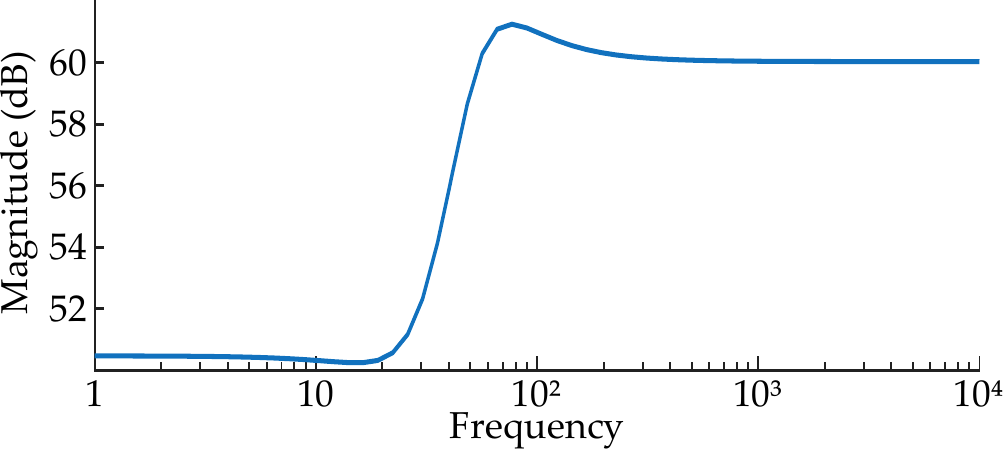}} \\
   \cmidrule{2-6}

   &
   \begin{tabular}{c@{\hspace{2pt}}c} \rotatebox[origin=c]{90}{Proportional} & \rotatebox[origin=c]{90}{Integral} \end{tabular}    & 
   % \begin{tabular}{c}$K(s) = k_1 + b_1s$, $D(s) = 0$ \\ $C_A(s) = 0$, $C_L(s) = k_d$\end{tabular} 
   \begin{tabular}{c} $C_A(s) = k_d + i_d/S$ \\ $C_L(s) = 0$\end{tabular} 
   &     $\dfrac{k_1(\alpha + k_d + i_d/s)}{\alpha + k_1 + k_d + i_d/s}$               &      \begin{tabular}{c@{}c}{\begin{tabular}{c}$bk_1^2\omega^2 - i_dk^2$\end{tabular}} & $\geq 0$\end{tabular}             &     \raisebox{-0.4\height}{\includegraphics[width=0.4\columnwidth]{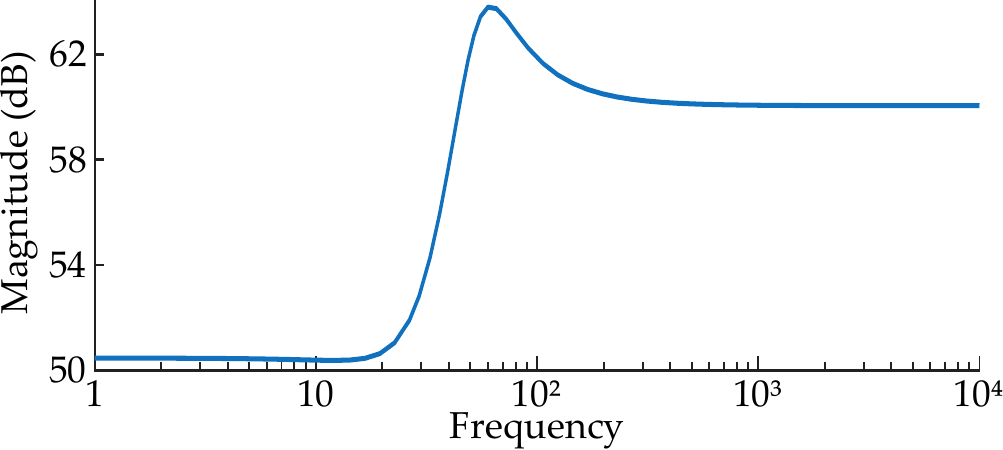}} \\
       \bottomrule
\end{tabular}
\vspace{-1.em}
\end{table*}

\subsubsection{Load-Side Control}
We first consider controlling the system from the load side ($C_A(s) = 0$). Here we place the encoder on the load, and use feedback from this encoder to modulate the actuator torque. 
Since we ultimately want to move the load to the desired position, using load-side control intuitively seems like the most direct way to reach accurate performance. 
However, as we will see from our analysis, this is not always the case.

\p{Proportional Control}
This setup replicates the \textit{Pure Spring} setup in Section \ref{ss:t1}.
The passivity analysis for this system is discussed in Section \ref{ss:t1}, and the frequency response is provided in Table \ref{tab:cont}.

\p{Proportional-Derivative Control}
We next consider a PD controller on the load side with $C_L(s) = k_d + b_ds$.
Row 1 in Table \ref{tab:cont} gives the stiffness transfer function and the passivity constraints for this controller.
These conditions suggest that this system will always be passive if $k_d\leq (b+b_d)k/b$ and $-b_dk_1m \geq 0$. While the first condition can be satisfied by tuning the values of $k_d$ and $b_d$, the second condition cannot be achieved for a compliant system with PD control ($b_d, k_1 > 0$). 
Thus, a PD controller on the load-side cannot be used for designing a safe SEA. 

\p{Proportional-Integral Control}
The final load-side method we consider is a PI controller ($C_L(s) = i_d/s + k_d$), where $i_d$ is the integral gain of this controller. 
The stiffness transfer function and the conditions for passivity are summarized in Row 2 of Table \ref{tab:cont}.
Similar to PD control, we cannot find proportional and integral gains of the system that would ensure passivity for all frequencies $\omega$. 
This suggests that --- for load-side control of a standard SEA --- the only feasible solution is a proportional controller with controller gains $k_d \leq k_1$. 
But this goes back to our original problem: the constraints on the controller gains and the lack of any physical or controller damping leads to oscillations in the system, and inherently limits its performance (see Section \ref{sec:sims}). 
Thus, we require a controller for which the gains can be tuned to minimize oscillations and maximize performance.

\subsubsection{Actuator-Side Control}
With this problem in mind, we finally consider actuator-side control for SEAs ($C_L(s) = 0$). 
These approaches track the actuator position and determine the actuator torques using $C_A$. 
Note that under this controller we have no direct feedback based on the position of the load, but can only \textit{indirectly} affect its position via the actuator.

\p{Proportional Control}
Similar to load-side control, we first start with proportional control on the actuator side ($C_A(s) = k_d$). 
The stiffness transfer function and the passivity conditions for this system --- summarized in Row 4 of Table \ref{tab:cont} --- suggest that the system will be passive for \textit{all values} of controller stiffness.
The frequency response for this system suggests that, as the frequency increases, the perceived stiffness of the system converges to the physical stiffness of the spring. 
Contrary to load-side control, this approach does not have any constraints on the controller gains, allowing us increase the controller gains to achieve a better performance while also maintaining the safety of the SEA system.
This is promising; but, as we will experimentally show, without any way to remove energy from the system the load-side tracking error is still significant.

\p{Proportional-Derivative Control}
We now consider a PD controller on the actuator-side ($C_A(s) = k_d + b_ds$). 
Using these values in \eq{m10}, we get the stiffness transfer function for the system as summarized in Row 5 of Table \ref{tab:cont}.
Similar to the proportional control, the passivity condition for this system in Table \ref{tab:cont} will always be satisfied for all values of the controller stiffness as the condition does not depend on $k_d$.
In our simulations, we leverage this controller and show that this system can ensure safety as well as accurate tracking of the actuator position. However, since there is no damping on the load side, we have no control over the oscillations of the load and thus compromise the performance.

\p{Proportional-Integral Control}
Finally, we test a actuator-side PI controller where ($C_A(s) = k_d + i_d/s$).
The last Row of Table \ref{tab:cont} outlines the stiffness transfer function and the passivity conditions for this controller.
Similar to the integral control on the load-side, since the physical spring stiffness of the system cannot be negative, we cannot ensure that the passivity condition is satisfied. 
From this combined analysis, we conclude that for the actuator-side control of series elastic actuator with pure spring, PD control is a feasible solution for safe and compliant interaction.
Further testing will be done in Section~\ref{sec:sims} to explore its performance.

\subsection{Combining Mechanical and Control Solutions} \label{ss:t3}
So far we have discussed separate solutions for mechanical properties and control parameters. 
We now consider a combined solution that factors in both the mechanical design of the system as well as its control methodology. 
Building on our observations from the previous sections, we propose a two part solution: i) placing a damper in the transmission to reduce oscillations between actuator and load while ii) using actuator-side PD control to guarantee passivity over a large range of values. 
We perform passivity analysis on this system and outline conditions under which it is passive.

Since we are performing actuator control with parallel spring-damper, we have $C_L(s) = 0$ and $D_2(s) = 0$.
The damped compliant element in the series elastic actuator can be represented using $D_1(s) = k_1 + b_1s$, and the controller on the actuator side can be written as $C_A(s) = k_d + b_ds$. 
Substituting these values in \eq{p4}, we write the stiffness transfer function of the system as:
\begin{equation} \label{eq:10}
    \dfrac{F_L}{X_L} = \dfrac{(k_1 + b_1s)(\alpha + b_ds + k_d)}{\alpha + (b_1 + b_d)s + k_1 + k_d}
\end{equation}
We leverage \eq{p7} along with the impedance transfer function for the system to get the passivity condition:
\begin{equation}
\begin{split}
    & m^2b_1\omega^6 + b_1((b+b_d)^2 +bb_1 + b_1b_d - 2mk_d)\omega^4\\ 
    & + (b_1k_d^2 + bk_1^2 + b_dk_1^2)\omega^2 
\end{split}
    \geq 0
\end{equation}
In this condition the coefficients of $\omega^6$ and $\omega^2$, are always positive. 
Eliminating these terms from the inequality, we can rewrite the simplified condition for passivity of this system as follows:
\begin{equation}
    b_1((b+b_d)^2 +bb_1 + b_1b_d - 2mk_d)\omega^4 \geq 0
\end{equation}
Here we observe that the passivity of the system \textit{does not depend} on the physical stiffness of the spring.
In other words --- we can have a soft transmission, and still maintain passivity with high control gains.
To ensure the passivity of the system, we only need the controller stiffness $k_d \leq ((b+b_d)^2 +bb_1 + b_1b_d)/2m$.
This is effectively a constraint on the ratio between the controller stiffness $k_d$ and the controller damping $b_d$.
We can ensure the passivity of the system by tuning the controller's gains, $k_d$ and  $b_d$, such that this condition is satisfied. 
Additionally, by tuning the value of the physical compliance $D_1$, we can reduce the mechanical oscillations in the system while also making the system compliant to any external impacts. 
The ability to independently tune the physical and controller parameters of the system enables simultaneous optimization for both safety and performance.

\section{Simulations} \label{sec:sims}

So far we have enumerated different mechanical and control solutions and performed theoretical analysis for the stability and compliance of those approaches. 
But what about system performance?
Restricting our focus to only the feasible options identified in Section~\ref{sec:theory}, we here conduct controlled simulations to empirically evaluate the performance of passive SEAs.

\p{Experimental Setup}
We consider a 1-DoF SEA with actuator mass $m = 1 kg$ and internal damping $b = 10 N\cdot s/m$. 
To simulate a robotic system where the actuator is connected to the robot link via a transmission, we connect the compliant element of the SEA $D_1 = k_1 + b_1s$ to a pure mass $m_L$.
Prior works have shown that pure masses are worst-case scenarios for SEA performance and stability \cite{losey2017effects,sergi2015stability}.
We therefore set $D_2 = 0$ to evaluate the coupled stability of the system. 

\begin{figure*}[]
    \centering
    \includegraphics[width=0.97\linewidth]{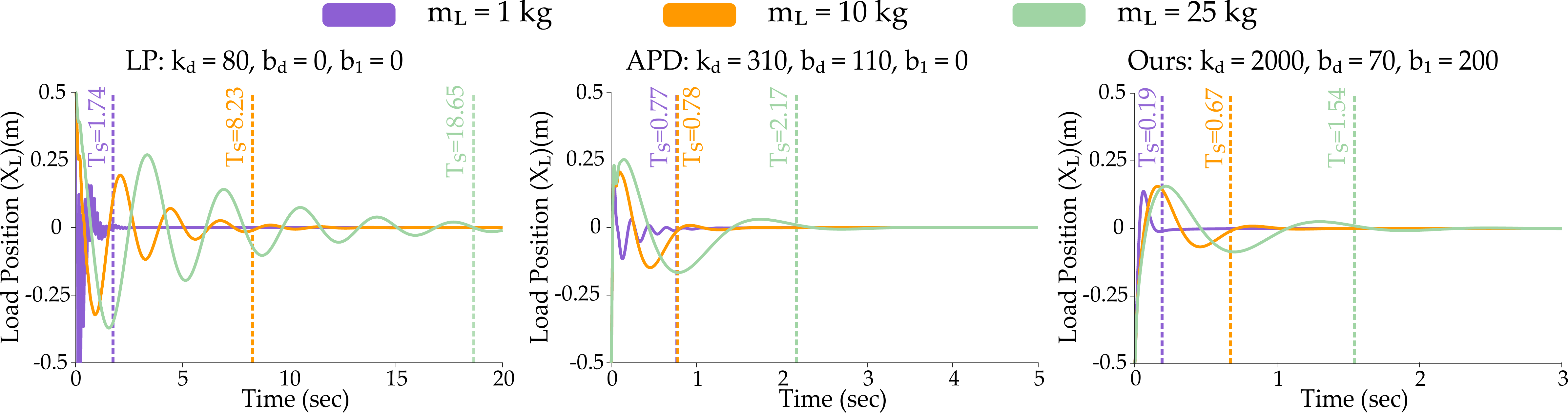}
 	\vspace{-0.5em}
    \caption{Results for simulated experiments. We evaluate the ability of the different controllers to bring the load to equilibrium across a variety of load masses. (Left) The performance of load-side proportional control. (Middle) The object position using actuator-side PD control. (Right) The performance of Ours. Across all mass values, Ours leads to a faster convergence with fewer oscillations in the system.}
    \label{fig:sims}
 	\vspace{-1.5em}
\end{figure*}

\p{Independent Variables}
We compare the performance of our proposed coupled mechanical and control solution (\textbf{Ours}) to two approaches discussed in Section \ref{sec:theory}: 1) Pure spring with \textbf{L}oad-side \textbf{P}roportional control (\textbf{LP}), and 2) Pure spring with \textbf{A}ctuator-side \textbf{PD} control (\textbf{APD}). 
For each of these approaches, we fix stiffness of the compliant element $k_1 = 1000N/m$ and the mass of the load ($m_L = 10 kg$), and compute the controller parameters $k_d$,  $b_d$ and $b_1$ that lead to the best case performance of each system (See Figure \ref{fig:sims} for optimized values).

While we compute these optimal parameters with a fixed load mass $m_L$, in a real world setting, the robot may need to move objects with varying masses. 
In such situations, a controller tuned for a fixed value of load mass $m_L$ should be able to perform well even when the mass of the load changes. 
Assuming the payload of the robot to be $25 kg$, we evaluate the performance of the controllers for three discrete mass values $m_L = \{1kg, 10kg, 25kg\}$.

\p{Dependent Variables}
To evaluate the performance of the system we provide a step input to the position of the load, and the controller tries to bring the system back to equilibrium position. 
For each controller and selected load mass we measure the final state error in the position of the load and the settling time required for the controller to converge within $2\%$ of the steady-state position. 

\p{Results}
Our results for this simulation are summarized in Figure \ref{fig:sims}.
We observe that for all controllers and load mass values the error in the load position converges to $0$ after being displaced by a step input of $0.5 m$. 
However, the time taken by the load to reach this equilibrium position varies for the different approaches.
For all mass values, Ours converges to the equilibrium in shorter time as compared to LP and APD. 
We also observe that LP and APD induce high frequency oscillations in the system for lower mass values, and LP induces low frequency oscillations even at higher loads. 
On the other hand, Ours converges to the equilibrium with fewer oscillations as compared to the baselines across all loads.

These results suggest that while LP can keep the system passive, it leads to a poor tracking of the load position. 
Similarly --- due to a lack of physical damping --- as the load increases, the performance of APD deteriorates as well. 
Our proposed approach for combining physical damping with actuator-side control enables accurate tracking of the load position and brings the system to equilibrium in a shorter time as compared to the baselines.
\section{Experiments}

\begin{figure}
    \centering
    \includegraphics[width=1.0\linewidth]{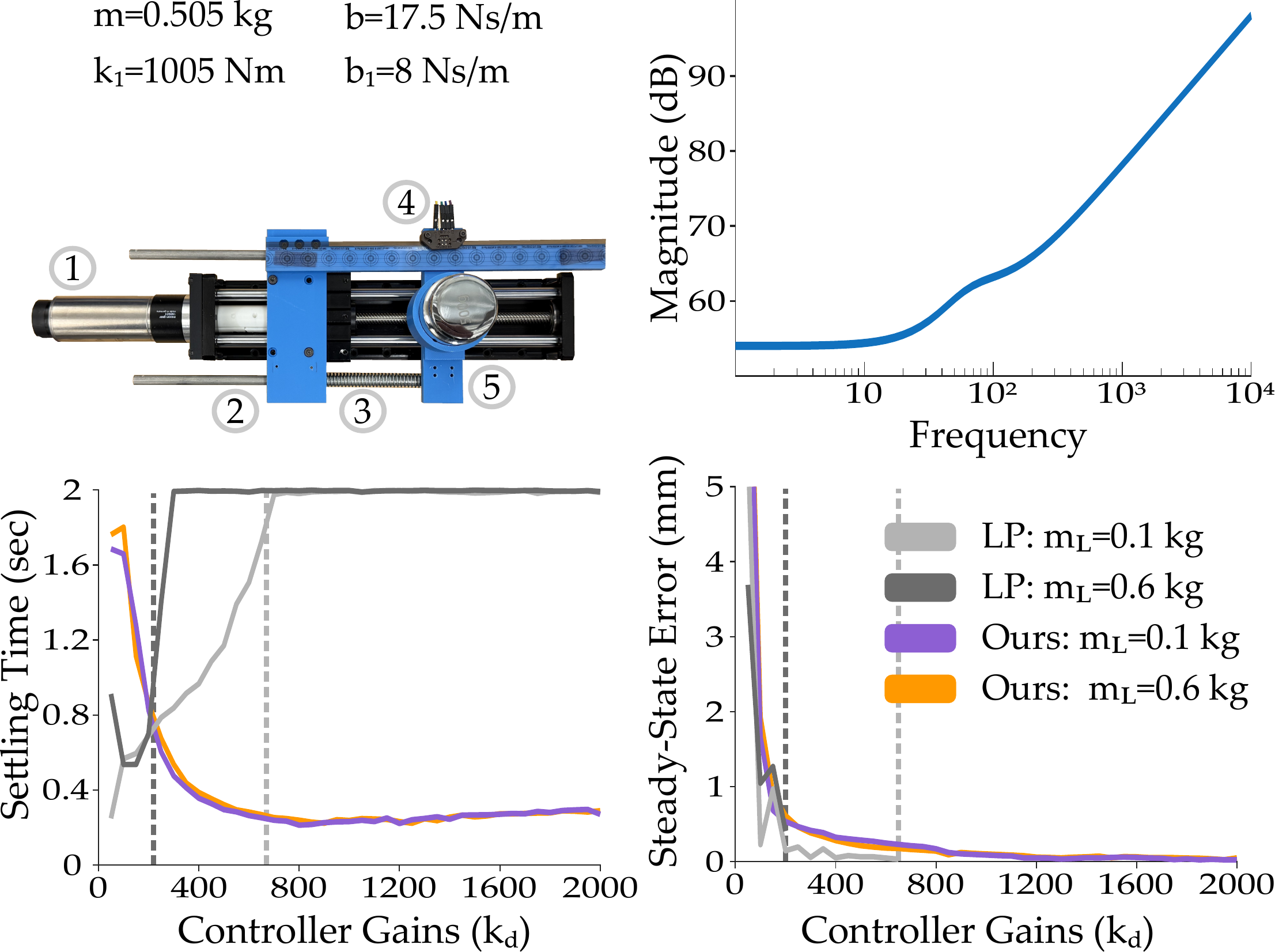}
 	\vspace{-1.0em}
    \caption{Results for real world experiments. (Top Left) Physical setup of the SEA with the following elements: (1) Actuator (2) Friction Damper (3) Spring (4) Linear Incremental Encoder (5) Output with load. (Top Right) Perceived stiffness for Ours during sudden impacts with an operator. The bottom row shows the settling time (left) and steady state error (right) of each controllers for different values of mass attached at the load. We observe that as the control gains increase, Ours leads to a better performance as compared to LP. The dotted lines in the figures highlight the controller gains after which LP becomes unstable.}
    \label{fig:exp}
 	\vspace{-0.9em}
\end{figure}

Now that we have validated the performance of our combined mechanical and control solution in controlled simulations, we finally test our approach and the baselines in the real world with a physical series elastic actuator (see top left image in \fig{exp}). Videos of our real-world experiments are available at: \href{https://youtu.be/jmJ3YBkuHo0}{https://youtu.be/jmJ3YBkuHo0}.

\p{Experimental Setup}
The physical setup for the experiment is shown in Figure \ref{fig:exp} (top left). 
The system is actuated using a Maxon Motor EC-30, with a gear ratio of $4.8$, connected to the load via a guide rail on a linear transmission. 
The position of the motor is tracked using a rotary incremental encoder (MR Encoder, 1000 cpt) and the position of the load with respect to the spring ($X_L - X_A$) is measured using a linear encoder (US-Digital EM1-0-500-I).
We compute the mass of the actuator ($m$), the actuator damping $b$, and the spring stiffness and damping ($k_1, b_1$) empirically and report the identified values of these parameters in Figure \ref{fig:exp}. 
The system was controlled by connecting the actuator and encoders to MATLAB and operated at a control frequency of $500$ Hz. 

\p{Procedure} 
In this experiment --- similar to our controlled simulations --- we compare the performance of our proposed approach (Ours) to a load-side proportional control (LP), which can ensure passivity of the system in the presence of a damper (see Section \ref{sec:theory}).
LP was chosen in part because it is the standard mechanical and control strategy.\cite{vallery2007passive,oh2016high,shardyko2021series}
We set the controller damping for the LP as $b_d=0$ (since LP is a proportional controller) and for our proposed approach as $b_d=25$. 
We evaluate the performance of the controllers paired with the damped physical system for different values of controller stiffness $k_d \in [0, 2000]$ and for two values of mass attached at the actuator output $mL = \{0.1, 0.6\}$.

We first analyze the frequency response of the stiffness transfer function for our approach given in \eq{10}. 
To see the effect of damping on the perceived stiffness in real-world situations, we compute the perceived stiffness of our system averaged over $10$ independent physical impacts using $k_{out} = F_L/X_L$. 
These impacts can be approximated as step changes in load force.
We then evaluate how quickly and accurately the system moves the load to a desired position after being displaced by an impulse. 
To test this condition, we initialize the load at the $20mm$ position on the guide rails and the motor tries to move that load to the equilibrium position.
For each value of the controller gain $k_d$ and load mass $mL$, we compute the time required by the controller to bring the system to equilibrium (settling time). 
Since robotic systems need to reach the desired position quickly after being displaced, we allow a maximum of $2\ sec$ for the controller to bring the system to equilibrium.
For the testing conditions where the controller settles to a constant value, we also compute the steady-state error in the position of the load.
Ideally, we want this error to be as close to $0$ as possible.

\p{Results}
The results for this experiment are summarized in Figure \ref{fig:exp}.
From the frequency response plot in \ref{fig:exp}, we observe that the perceived stiffness of the system increases with an increase in frequency. 
The observed stiffness of the system in the real setup --- averaged across $10$ separate impacts --- was observed to be $k_{out} = 2292 \pm 91.6 N/m$, which is greater than the physical stiffness of the system $k_1 = 1005N/m$. This shows that while the perceived stiffness of the system increases because of the mechanical damping in the transmission, the system is still compliant and can soften the force of impact on collision. 
Additionally, by choosing more compliant physical springs, systems with lower perceived stiffness can be achieved. 

Now that we have analyzed the safety and compliance characteristics of our approach, we move on to evaluate the performance of the system. 
We observe that, in this damped system, as the controller gains are increased, Ours leads to a lower settling time as compared to the LP.
We also observe that as the controller gains are increases, LP eventually leads to unstable system behavior. 
Due to the physical and controller damping, our approach has a higher error in tracking the load position for lower values of controller gains $k_d < 200$.
However, as the controller gains are increased, these errors converge to $0$, highlighting the ability of our proposed approach to accurately control the system.

\section{Conclusion}

We propose a mechanical and control SEA configuration that enables accurate tracking while ensuring safety during human-robot interaction. 
We consider 1-DoF SEAs and start by analyzing the passivity and compliance of different mechanical designs and control solutions. 
Based on this analysis, we propose the use of a physically damped SEA paired with a proportional-derivative controller on the actuator side. 
Our theoretical analysis and experiments demonstrate that actuator-side control enables a wider range of control gains than previously possible, and the mechanical damper combines with those gains to minimize oscillations and maximize performance.
This result fundamentally advances the design and control of SEAs; future works can develop more complex actuator-side controllers that build on our findings.

%%%%%%%%%%%%%%%%%%%%%%%%%%%%%%%%%%%%%%%%%%%%%%%%%%%%%%%%%%%%%%%%%%%%%%%%%%%%%%%%%

% \newpage
% \vspace{-0.5em}
\balance
\bibliographystyle{IEEEtran}
\bibliography{IEEEabrv,bibtex}

\end{document}